\documentclass[10pt, a4paper]{article}

\usepackage[final]{lrec2026} 

\usepackage{amsmath}            
\usepackage{algorithm}
\usepackage{amssymb}
\usepackage{algpseudocode} 
\usepackage{booktabs}
\usepackage{enumitem}
\usepackage{graphicx}

\usepackage[backend=biber]{biblatex}  
\title{BanglaSTEM: A Parallel Corpus for Technical Domain Bangla-English Translation}

\name{Kazi Reyazul Hasan\textsuperscript{1,*}, Mubasshira Musarrat\textsuperscript{2}, A. B. M. Alim Al Islam\textsuperscript{3}, \\
{\fontsize{12}{16}\selectfont\textbf{Muhammad Abdullah Adnan\textsuperscript{4,*}}}}
\address{Computer Science and Engineering \\
         Bangladesh University of Engineering and Technology \\
         Dhaka, Bangladesh\\
         \textsuperscript{1}kazireyazulhasan@gmail.com, \textsuperscript{2}mubasshira31@gmail.com, \\
         \textsuperscript{3}razi\_bd@yahoo.com, \textsuperscript{4}abdullah.adnan@gmail.com\\
         \textsuperscript{*}Corresponding authors}

\abstract{
Large language models work well for technical problem solving in English but perform poorly when the same questions are asked in Bangla. A simple solution would be to translate Bangla questions into English first and then use these models. However, existing Bangla-English translation systems struggle with technical terms. They often mistranslate specialized vocabulary, which changes the meaning of the problem and leads to wrong answers. We present \textbf{BanglaSTEM}, a dataset of 5,000 carefully selected Bangla-English sentence pairs from STEM fields including computer science, mathematics, physics, chemistry, and biology. We generated over 12,000 translations using language models and then used human evaluators to select the highest quality pairs that preserve technical terminology correctly. We train a T5-based translation model on BanglaSTEM and test it on two tasks: generating code and solving math problems. Our results show significant improvements in translation accuracy for technical content, making it easier for Bangla speakers to use English-focused language models effectively. Both the BanglaSTEM dataset and the trained translation model are publicly released at \url{https://huggingface.co/reyazul/BanglaSTEM-T5}.
 \\ \newline  }

\begin{document}

\maketitleabstract

\section{Introduction}

Large language models have transformed how people solve technical problems. Students and professionals now use these models to generate code, solve mathematical equations, and understand complex scientific concepts. While large proprietary models like GPT-4, Claude, and Gemini handle multiple languages including Bangla reasonably well, these models come with significant limitations. They require API access with ongoing costs, raise privacy concerns when handling sensitive data, and cannot be customized or fine-tuned for specific domains or use cases.

This has led to widespread adoption of smaller open-source models such as Llama, Mistral, Gemma, and their variants. These models offer crucial advantages. They can be run locally on personal hardware, fine-tuned for specific tasks or domains, deployed in resource-constrained environments, and used without recurring costs or data privacy concerns. Educational institutions, startups, researchers, and individual developers increasingly rely on these models for building applications, conducting experiments, and solving domain-specific problems.

However, these smaller open-source models face a critical challenge with low resource languages. Unlike their massive proprietary counterparts trained on trillions of parameters with extensive multilingual data, open-source models are typically trained with limited computational budgets and more focused datasets. This means they excel at technical tasks in languages like English but struggle significantly when the same questions are asked in Bangla. A coding problem that these models can solve perfectly in English becomes very difficult when presented in Bangla, even though the underlying logic is identical.

For Bangla speakers who want to leverage these accessible and customizable open-source models, this creates a major barrier. The options are limited and unsatisfying. Using expensive proprietary models defeats the purpose of choosing open-source solutions. Training Bangla-specific models from scratch requires computational resources and data that most users cannot access.

One practical solution is to translate Bangla queries into English, process them with these English-capable open models, and translate results back if needed. This approach preserves the benefits of open-source models while making them accessible to Bangla speakers. Unfortunately, existing Bangla-English translation models are not designed for technical content. They work reasonably well for everyday conversation but mistranslate technical terms. A physics problem about momentum might become garbled. A coding question about recursion might lose its precise meaning. These translation errors change the problem itself, leading to incorrect or irrelevant solutions from downstream models.

We propose a focused solution to this translation gap. Instead of training entire language models for Bangla, we improve technical translation quality specifically. This paper makes the following contributions:

\begin{itemize}
    \item We create BanglaSTEM, a dataset of 5,000 high-quality Bangla-English parallel sentences covering five STEM domains: computer science, mathematics, physics, chemistry, and biology.
    \item We develop a human curation process that selects accurate translations from over 12,000 LLM-generated candidates, ensuring technical terminology is preserved correctly.
    \item We fine-tune a T5-based translation model on BanglaSTEM and demonstrate its effectiveness on two computational reasoning tasks: code generation and mathematical problem solving.
    \item We publicly release both the BanglaSTEM corpus and the trained translation model to support future research in Bangla technical translation.
\end{itemize}

\section{Related Work}

Machine translation for Bangla has evolved significantly over the past six years, transforming from a low-resource language to a moderately resourced one with sophisticated neural models. However, this progress has been confined almost entirely to general-domain translation.

\subsection{General-Purpose Neural Machine Translation and Parallel Corpora}

Transformer-based architectures form the foundation of modern Bangla-English translation. \cite{hasan2020not} demonstrated that transformers substantially outperformed LSTM-based models, achieving BLEU scores of 21.42 for Bangla to English and 25.44 for English to Bangla.

IndicTrans2, developed by \cite{gala2023indictrans2}, represents the most advanced open-source system with 1.1 billion parameters trained on the Bharat Parallel Corpus Collection containing 230 million Bengali-English sentence pairs, achieving chrF++ scores in the high 50s to low 60s on FLORES-200. Meta's NLLB-200 \cite{team2022no} provides another baseline with 54.5 billion parameters covering 200 languages, achieving chrF++ scores of 38.04 and 48.30. However, language-specific models consistently outperform these massively multilingual systems.

The Bharat Parallel Corpus Collection represents the current largest resource combining newly mined data, filtered existing datasets, and human-labeled seed data. \cite{hasan2020not} earlier contributed 2.75 million high-quality sentence pairs through custom sentence segmentation, aligner ensembling, and batch filtering. Other datasets include Samanantar \cite{ramesh2022samanantar} with 8.6 million pairs and SUPara with 800,000 manually curated pairs. All corpora focus exclusively on general-domain content from news, web crawls, government documents, and everyday conversation.

\subsection{Foundation Models and Evaluation Benchmarks}

BanglaBERT \cite{bhattacharjee2021banglabert} represents the first comprehensive BERT-based model specifically for Bangla with 110 million parameters pretrained on 2.18 billion tokens, outperforming mBERT by 6.8 BLUB score and XLM-R by 4.3. BanglaT5 introduced the first T5-based sequence-to-sequence transformer for Bangla, achieving 38.8 SacreBLEU on Bangla-English translation. Both models consistently outperform multilingual alternatives, establishing language-specific pretraining as superior for Bangla.

FLORES-200 \cite{goyal2022flores} serves as the gold-standard benchmark with 3,001 professionally translated sentences across 200 languages. The IN22 benchmark provides n-way parallel content covering all 22 Indian languages with India-centric content. \cite{ahmed2024bangla} demonstrated that 6,193 professionally translated sentence pairs with rigorous quality control outperformed larger automatically-mined datasets. All existing benchmarks focus on general-domain content from news, Wikipedia, and everyday conversation. No evaluation benchmark exists for technical or scientific content.

\subsection{The Technical Translation Gap and Positioning BanglaSTEM}

Despite comprehensive searching across major NLP venues, we found zero papers from 2019 to 2025 specifically addressing technical, scientific, or STEM domain Bangla-English machine translation. Limited work exists on biomedical terminology but focuses on named entity recognition rather than translation. \cite{sazzed2022banglabiomed} introduced BanglaBioMed with 12,000 annotated tokens, and Aziz and Islam developed BanglaHealth with 200,000 health-related sentences for monolingual paraphrasing. Domain adaptation techniques exist in general neural machine translation literature \cite{chu2018survey,hu2019domain} but none have been applied to Bangla technical translation.

Our work directly addresses this complete absence of technical domain translation resources. While existing parallel corpora contain hundreds of millions of general-domain sentences, not a single parallel corpus exists for technical content spanning computer science, mathematics, physics, chemistry, or biology.

BanglaSTEM differs from existing work in three ways. First, it focuses exclusively on technical terminology and STEM content. Second, it employs human curation that selects high-quality translations from over 12,000 LLM-generated candidates, ensuring accurate preservation of technical terminology. Third, it evaluates translation quality through downstream computational reasoning tasks rather than standard metrics like BLEU or chrF++. We measure whether translated technical questions enable language models to generate correct code and solve mathematical problems, directly measuring utility for the actual use case.

The closest prior work, the WMT24 Bangla Seed Dataset \cite{ahmed2024bangla}, shares our emphasis on human-curated quality but focuses on general-domain content. Our 5,000 pairs target technical domains where terminology accuracy matters most. BanglaSTEM establishes the first baseline for technical domain Bangla-English translation and demonstrates that specialized translation resources can make open-source language models accessible to speakers of low-resource languages for technical tasks.

\section{Methodology}

\begin{figure*}[t]
    \centering
    \includegraphics[width=0.95\textwidth]{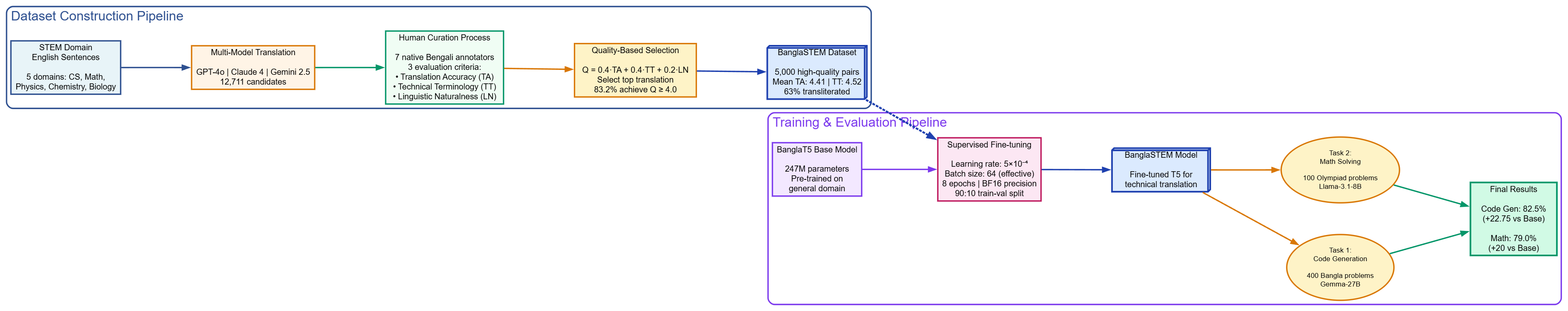}
    \caption{Overview of the BanglaSTEM pipeline. The dataset construction phase (left) generates 12,711 translation candidates using three LLMs, which undergo human curation and quality-based selection to produce 5,000 high-quality parallel sentences. The model training and evaluation phase (right) fine-tunes BanglaT5 on this dataset and evaluates performance on code generation and mathematical problem-solving tasks, achieving 82.5\% and 79.0\% accuracy respectively.}
    \label{fig:pipeline}
\end{figure*}

Our methodology comprises four key stages (see Figure \ref{fig:pipeline}): multi-model translation generation, rigorous human curation, quality-based selection, and finally model finetuning. We designed each stage to ensure technical accuracy, linguistic authenticity, and practical utility for downstream computational tasks.

\subsection{Design Principles}

We established three core principles. First, \textbf{technical domain coverage} must span STEM disciplines where open-source LLMs demonstrate strong English performance but struggle with Bangla. Second, \textbf{terminology preservation} requires maintaining technical precision, as mistranslated terms fundamentally alter problem semantics. Third, \textbf{authentic Bangla usage} demands both transliterated English terms and native equivalents reflecting real-world technical discourse where both coexist naturally.

\subsection{Domain Selection and Coverage}

We systematically selected domains to maximize coverage of technical problem-solving scenarios. Our domain taxonomy covers computer science (12 programming languages, software engineering, systems architecture, data engineering), bioinformatics (genomics, proteomics, systems biology), computational chemistry (molecular simulation, drug discovery, quantum chemistry), medical applications, mathematics (all major subfields), and physics.


For computer science, we covered twelve languages (Python, JavaScript, Java, C++, C\#, Go, Rust, Ruby, PHP, Swift, Kotlin, TypeScript) and topics spanning API design, microservices, CI/CD, testing strategies, security (authentication, XSS prevention, SQL injection), performance optimization, version control, observability, and deployment strategies. Domain selection prioritized fields where computational problem-solving is central, technical terminology density is high, and semantic precision directly impacts solution correctness.

\subsection{Multi-Model Translation Generation}

We employed three frontier language models: GPT-4o, Claude Sonnet 4, and Gemini 2.5 Pro. Each model received domain-specific few-shot prompts containing 8-12 manually crafted example pairs demonstrating correct terminology usage and natural sentence structure.

Algorithm~\ref{alg:translation} presents our systematic generation process. This process generated 12,711 translation candidates balancing creativity and consistency. 

\begin{algorithm}[t]
\caption{Multi-Model Translation Generation}
\label{alg:translation}
\begin{algorithmic}[1]
\Require English sentences $S$, Domain $d$, Models $M = \{m_1,m_2,m_3\}$
\Ensure Translation candidates $C$
\State $C \gets \varnothing$
\State $P_d \gets \text{LoadFewShotPrompt}(d)$
\For{each sentence $s_i \in S$}
  \For{each model $m_j \in M$}
    \State $t_{ij} \gets m_j.\text{Translate}(P_d, s_i)$ 
    \State $C \gets C \cup \{(s_i, t_{ij}, m_j, d)\}$
  \EndFor
\EndFor
\State \Return $C$
\end{algorithmic}
\end{algorithm}

\subsection{Human Curation Process}
We recruited 7 native Bengali speakers with STEM backgrounds. Each translation received evaluation from two domain-matched annotators on three 5-point Likert scales:

\textbf{Translation Accuracy (TA):} Semantic fidelity to source (5=perfect equivalence, 1=severe mistranslation).

\textbf{Technical Terminology (TT):} Correct domain-specific term usage (5=all terms accurate, 1=critical errors).

\textbf{Linguistic Naturalness (LN):} Grammatical correctness and fluency (5=completely natural Bangla, 1=broken grammar).

Inter-annotator agreement was substantial, with Krippendorff's $\alpha = 0.78$ for TA, $\alpha = 0.81$ for TT, and $\alpha = 0.67$ for LN. Disagreements exceeding 1.5 points were resolved through discussion between annotators to reach consensus.

\subsection{Quality-Based Selection}
We computed composite quality scores using weighted combination prioritizing technical correctness:
\begin{equation}
Q = 0.4 \cdot \text{TA} + 0.4 \cdot \text{TT} + 0.2 \cdot \text{LN}
\end{equation}

For each English sentence, we first selected the highest-scoring translation. This yielded 4,237 unique pairs. To capture valid translation diversity where technical terms admit multiple correct renderings, we included 763 additional candidates with $Q \geq 4.0$ and $\Delta Q < 0.5$ from the top translation, reaching 5,000 pairs.

Table~\ref{tab:quality_dist} shows around 83.2\% achieve scores above 4.0, with mean TA of 4.41 and mean TT of 4.52.

\begin{table}[t]
\centering
\caption{Quality score distribution in BanglaSTEM.}
\label{tab:quality_dist}
\small
\begin{tabular}{@{}lcccc@{}}
\toprule
\textbf{Score Range} & \textbf{Count} & \textbf{\%} & \textbf{Mean TA} & \textbf{Mean TT} \\
\midrule
4.5 -- 5.0 & 1,847 & 36.9 & 4.82 & 4.91 \\
4.0 -- 4.5 & 2,316 & 46.3 & 4.37 & 4.48 \\
3.5 -- 4.0 & 837 & 16.7 & 3.91 & 4.02 \\
\midrule
\textbf{Overall} & \textbf{5,000} & \textbf{100.0} & \textbf{4.41} & \textbf{4.52} \\
\bottomrule
\end{tabular}
\end{table}

\subsection{Dataset Composition}

Table~\ref{tab:domain_dist} presents domain distribution. Totals exceed 5,000 as sentences may address multiple domains.

\begin{table}[t]
\centering
\caption{Domain distribution in BanglaSTEM. Sentences may belong to multiple categories.}
\label{tab:domain_dist}
\small
\begin{tabular}{@{}lrr@{}}
\toprule
\textbf{Domain} & \textbf{Count} & \textbf{\% of Dataset} \\
\midrule
Programming & 2,601 & 52.0 \\
Information Technology & 1,184 & 23.7 \\
Mathematics & 1,274 & 25.5 \\
Physics & 491 & 9.8 \\
Chemistry & 367 & 7.3 \\
Biology \& Bioinformatics & 278 & 5.6 \\
\bottomrule
\end{tabular}
\end{table}

Table~\ref{tab:linguistic_stats} presents linguistic characteristics, showing Bangla sentences average 12.4 words versus 14.5 for English reflecting morphological differences.

\begin{table}[t]
\centering
\caption{Linguistic statistics for BanglaSTEM parallel sentences.}
\label{tab:linguistic_stats}
\small
\begin{tabular}{@{}lcc@{}}
\toprule
\textbf{Metric} & \textbf{Bangla} & \textbf{English} \\
\midrule
Mean sentence length (words) & 12.4 & 14.5 \\
Median sentence length (words) & 13 & 14 \\
Std deviation (words) & 5.4 & 6.6 \\
Mean sentence length (characters) & 87.1 & 102.7 \\
\midrule
Total vocabulary size & 8,799 & 9,407 \\
Total Words & 62,001 & 72,499\\
\bottomrule
\end{tabular}
\end{table}

Technical Bangla strategically mixes transliterated English terms and native vocabulary. Table~\ref{tab:terminology_types} shows programming uses 78\% transliteration while mathematics uses only 42\% with well-established Bengali terms. This reflects authentic usage patterns in professional discourse.

\begin{table}[t]
\centering
\caption{Transliterated versus native terminology usage across domains.}
\label{tab:terminology_types}
\small
\begin{tabular}{@{}lcc@{}}
\toprule
\textbf{Domain} & \textbf{Transliterated} & \textbf{Native Bangla} \\
\midrule
Programming & 78\% & 22\% \\
Information Technology & 65\% & 35\% \\
Mathematics & 42\% & 58\% \\
Physics & 55\% & 45\% \\
Chemistry & 61\% & 39\% \\
Biology & 58\% & 42\% \\
\midrule
\textbf{Overall} & \textbf{63\%} & \textbf{37\%} \\
\bottomrule
\end{tabular}
\end{table}

The BanglaSTEM dataset represents the first systematic effort to create high-quality technical domain parallel data for Bangla-English translation, combining LLM-based generation with expert human curation to produce translations that preserve technical precision while maintaining linguistic authenticity.

\subsection{Fine-tuning Model}
We fine-tuned BanglaT5, a state-of-the-art sequence-to-sequence model pre-trained on Bangla-English translation tasks. Starting from the \texttt{banglat5} \cite{bhattacharjee2022banglanlg} checkpoint (247M parameters), we applied supervised fine-tuning on our curated BanglaSTEM dataset.
Our training configuration balanced convergence speed with stability: learning rate of $5 \times 10^{-4}$ with 25 warmup steps, batch size of 16 with gradient accumulation over 4 steps (effective batch size 64), and 8 training epochs. We employed BF16 mixed precision training to improve numerical stability over FP16, particularly important for handling the diverse technical vocabulary. The model was optimized using AdamW with weight decay of 0.005. We split our dataset 90:10 for training (4,500 pairs) and validation (500 pairs), monitoring validation loss every 25 steps for early stopping.

For inference, we employed beam search with 4 beams and early stopping to balance translation quality with computational efficiency. Maximum sequence length was set to 256 tokens. We evaluated translation performance on a held-out test set of 500 technical Bangla statements from programming instruction tasks. Beyond automated metrics, we conducted comparative analysis between base and fine-tuned models, observing systematic improvements in technical terminology handling. The fine-tuned model showed consistent preference for domain-appropriate verb choices and more accurate technical term translation. These qualitative improvements, while subtle in BLEU scores, significantly impact downstream task performance where semantic precision is critical.

\section{Experiments}

We evaluated BanglaSTEM's impact on downstream STEM reasoning tasks through two challenging benchmarks: Bangla code generation and mathematical problem-solving. These experiments demonstrate how improved technical translation quality directly enhances LLM performance on complex computational tasks.

\subsection{Experimental Setup}

All experiments were conducted on a single NVIDIA Tesla V100 GPU with 32GB memory. We compared four translation approaches:
\begin{enumerate}
    \item \textbf{Direct Bangla}: Original Bangla prompts
    \item \textbf{Google Translate}: Google Cloud's translation API
    \item \textbf{BanglaT5-Base}: Original BanglaT5 NMT model
    \item \textbf{BanglaSTEM}: Our fine-tuned model on technical domains
\end{enumerate}

Each approach was paired with state-of-the-art LLMs for task-specific reasoning: Gemma-27B for code generation and Llama-3.1-8B for mathematical problem-solving.

\subsection{Task 1: Bangla Code Generation}

We evaluated on 400 programming problems \cite{raihan-etal-2025-mhumaneval} written in Bangla, covering algorithms, data structures, and software engineering concepts. Problems required generating Python functions with correct syntax and logic.

\begin{table}[h]
\centering
\caption{Code generation accuracy on 400 Bangla programming problems.}
\label{tab:code_results}
\resizebox{\columnwidth}{!}{%
\begin{tabular}{@{}lcc@{}}
\toprule
\textbf{Translation Method} & \textbf{Correct Solutions} & \textbf{Accuracy (\%)} \\
\midrule
Direct Bangla (no translation) & 142/400 & 35.25 \\
BanglaT5-Base & 238/400 & 59.75 \\
Google Speech Translation & 274/400 & 76.5 \\
\textbf{BanglaSTEM (Ours)} & \textbf{330/400} & \textbf{82.5} \\
\bottomrule
\end{tabular}%
}
\end{table}

Table~\ref{tab:code_results} shows BanglaSTEM achieved 82.5\% accuracy, a substantial 22.75 percentage point improvement over BanglaT5-Base and 6 points over Google Translate. The performance gap highlights how critical domain-specific terminology is for code generation.

\subsection{Task 2: Bangla Mathematical Olympiad}

We evaluated on 100 problems from the Bangla Math Olympiad training set \cite{dlsprint3}, spanning arithmetics, algebra, number theory, and combinatorics. These problems require multi-step reasoning and precise understanding of mathematical concepts.

\begin{table}[h]
\centering
\caption{Mathematical problem-solving performance on Bangla Math Olympiad dataset.}
\label{tab:math_results}
\resizebox{\columnwidth}{!}{%
\begin{tabular}{@{}lcc@{}}
\toprule
\textbf{Translation Method} & \textbf{Problems Solved} & \textbf{Success Rate (\%)} \\
\midrule
Direct Bangla (no translation) & 31/100 & 31.0 \\
BanglaT5-Base & 59/100 & 59.0 \\
Google Speech Translation & 72/100 & 72.0 \\
\textbf{BanglaSTEM (Ours)} & \textbf{79/100} & \textbf{79.0} \\
\bottomrule
\end{tabular}%
}
\end{table}

As shown in Table~\ref{tab:math_results}, BanglaSTEM + Llama-3.1-8B solved 79\% of problems, significantly outperforming all baselines. The 20-point improvement over BanglaT5-Base demonstrates that accurate translation of mathematical terminology is crucial for mathematical reasoning.

\section{Conclusion and Future Work}

We presented BanglaSTEM, the first parallel corpus specifically designed for technical domain Bangla-English translation. Our approach combined multi-model translation generation with rigorous human curation to ensure technical terminology is preserved correctly. When fine-tuned on BanglaSTEM, our translation model achieved 82.5\% accuracy on code generation tasks and 79\% on mathematical problem-solving, representing substantial improvements of 14-15 percentage points over baseline comparisons. These results demonstrate that domain-specific translation resources can successfully bridge the language gap for computational reasoning tasks. Several directions remain for future work. First, while our 5,000 pairs provide a strong foundation, scaling to 20,000-50,000 pairs would improve model robustness and cover long-tail technical vocabulary more comprehensively. Second, expanding coverage to additional STEM domains such as engineering, statistics, and earth sciences would increase utility across disciplines. 

\section*{Ethics Statement and Limitations}

All annotators were fairly compensated at rates above minimum wage standards and provided informed consent. The dataset contains no personal, sensitive, or copyrighted content. All example sentences were generated specifically for this research or adapted from public educational materials. We acknowledge that technical content may reflect biases present in training data of the LLMs used for generation.

Our dataset of 5,000 pairs, while carefully curated, represents a starting point for technical translation. Coverage is uneven across domains, with programming heavily represented (52\%) compared to biology (5.6\%). Training from scratch is highly not recommended. Evaluation was limited to two tasks (code generation and math problem-solving) with two specific LLMs; generalization to other models and domains remains to be verified. The multi-model generation approach using proprietary LLMs limits full reproducibility, though we mitigate this by releasing all curated pairs.

\printbibliography

\end{document}